\documentclass{article}
\usepackage{styles/spconf,amsmath,graphicx}
\usepackage{graphicx}
\graphicspath{{figures}}
\usepackage{todonotes}
\usepackage{hyperref}
\usepackage{booktabs}
\usepackage{subcaption}
\usepackage{amsfonts}

\def\cX{\mathcal{X}}
\def\cY{\mathcal{Y}}
\def\cF{\mathcal{F}}

\def\ea{\emph{et al}.\ }
\def\ni{\noindent}

\title{PU-EdgeFormer: Edge Transformer for Dense Prediction in Point Cloud Upsampling}
\name{Dohoon Kim\textsuperscript{1}, Minwoo Shin\textsuperscript{1}, and Joonki Paik\textsuperscript{1,2}
\thanks{This work was supported partly by Institute of Information \& communications Technology Planning \& Evaluation (IITP) grant funded by the Korea government (MSIT) (2021-0-01341, Artificial Intelligence Graduate School Program (Chung-Ang University))
and Field-oriented Technology Development Project for Customs Administration through National Research Foundation of Korea (NRF) funded by the Ministry of Science \& ICT and Korea Customs Service (2021M3I1A1097911).}}
\address{\textsuperscript{1}Department of Image, Chung-Ang University, Seoul, Korea \\
	\textsuperscript{2}Department of Artificial Intelligence, Chung-Ang University, Seoul, Korea}

\begin{document}

	\maketitle

	\begin{abstract}
		Despite the recent development of deep learning-based point cloud upsampling, most MLP-based point cloud upsampling methods have limitations in that it is difficult to train the local and global structure of the point cloud at the same time.
		To solve this problem, we present a combined graph convolution and transformer for point cloud upsampling, denoted by PU-EdgeFormer.
		The proposed method constructs EdgeFormer unit that consists of graph convolution and multi-head self-attention modules.
            We employ graph convolution using EdgeConv, which learns the local geometry and global structure of point cloud better than existing point-to-feature method.
		Through in-depth experiments, we confirmed that the proposed method has better point cloud upsampling performance than the existing state-of-the-art method in both subjective and objective aspects.
		The code is available at \url{https://github.com/dohoon2045/PU-EdgeFormer}.
	\end{abstract}
	\begin{keywords}
		Point cloud upsampling, neural networks, graph convolution, vision transformer
	\end{keywords}
	\section{Introduction}
	\label{sec:intro}
	Point clouds are the most widely used representation of three-dimensional (3D) data acquired with 3D sensors because they use less memory and computational cost than meshes or voxels.
	In recent research, point cloud is used as input to various 3D applications such as autonomous driving, 3D reconstruction, virtual/augmented reality, and robotics.
	However, the acquired point cloud is spatially sparse, and the coordinates are noisy and non-uniform because of various 3D acquisition problems including: i)~occlusion, ii)~light reflection, iii)~limited hardware, and iv)~computational cost. 
	When this point cloud is applied to 3D applications related to Neural Network as it is, information cannot be clustered due to sparsity and non-uniform characteristics.
	Therefore, it is very important to pre-process the raw point cloud because it causes performance degradation by failing to learn the contextual manifold of the local structure.
	Successful point cloud upsampling is an important preprocessing process that can improve the performance of various 3D tasks because it generates dense, uniform, and noise-free point clouds.
	
	Early point cloud upsampling used optimization methods~\cite{alexa2003computing, lipman2007parameterization}.
	After Qi \ea proposed PointNet, which successfully applied the deep learning method to the point cloud~\cite{qi2017pointnet}, Yu \ea first proposed deep learning-based point cloud upsampling method called PU-Net~\cite{yu2018pu}.
	Based on \cite{qi2017pointnet,yu2018pu}, Yu \ea proposed EC-Net to perform upsampling while preserving the edge of the point cloud~\cite{yu2018ec}.
	Yifan \ea proposed MPU upsampling progressively for each point patch~\cite{yifan2019patch}, and Li \ea proposed PU-GAN to which GAN and attention unit were applied~\cite{li2019pu}.
	Qian \ea proposed PUGeo-Net, which converts samples in 2D domain into 3D by applying linear transformation~\cite{qian2020pugeo}, and Qian \ea suggested NodeShuffle to the point shuffle process of the upsampling module to improve the performance of PU-GCN~\cite{qian2021pu}.
	
	Although the performance of deep learning-based point cloud upsampling is continuously improving, the 2D convolution operation used for feature extraction in most upsampling methods does not reflect the geometric relationship of the input point cloud because each point is learned independently.
	To solve this problem, MPU and PU-GCN have proposed methods of reflecting the local geometry of the input point cloud as a point feature, but there is still a problem that the global structure information of the input point cloud is not reflected as a point feature.
	
	In this paper, we propose a novel unit called EdgeFormer that extracts features from point cloud for successful upsampling.
	Using EdgeFormer we calculate the attention score from the input point through the multi-head self-attention of transformer~\cite{vaswani2017attention}, and designed the network to reflect the global structure in the point feature and also the local geometric structure through EdgeConv~\cite{wang2019dynamic}.
	
	\begin{figure*}[htb!]
		\centering
		\includegraphics[width=0.99\textwidth]{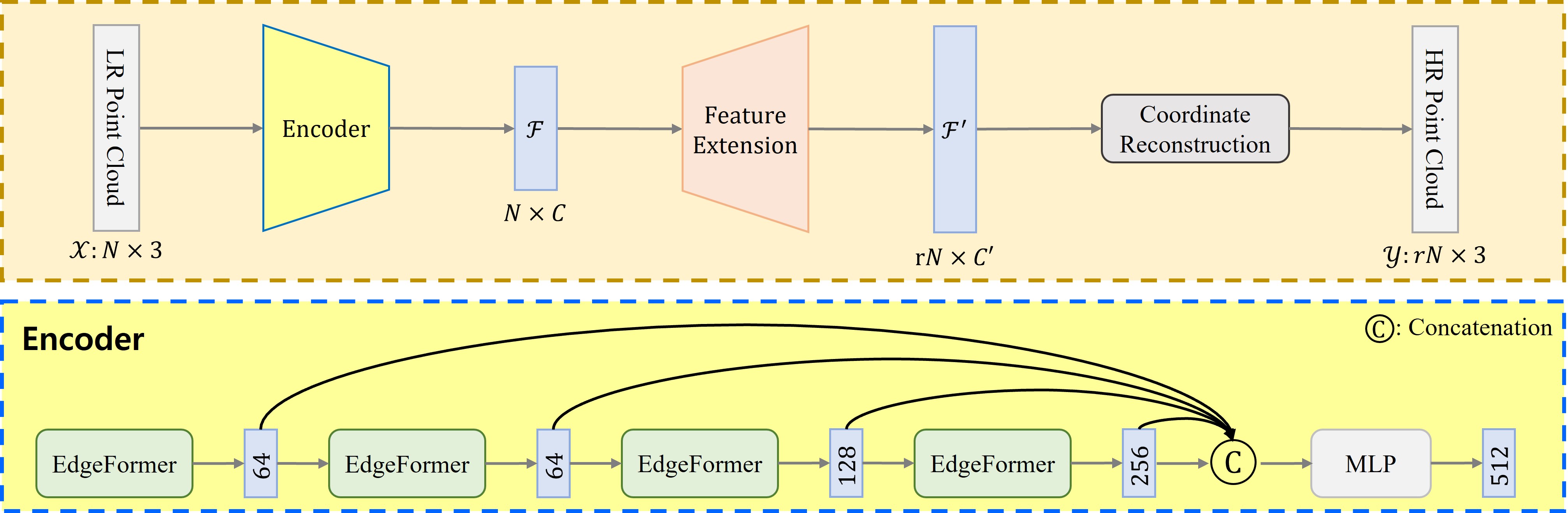}
		\caption{Architecture of PU-EdgeFormer~(top) and the encoder module~(bottom)}
		\label{fig:proposed}
	\end{figure*}
	
	\section{Proposed Method}\label{sec:proposed}
	Given a point cloud of $ N $ points, the point cloud upsampling process with an upsampling ratio $ r $ produces a dense point cloud $ \cY \in \mathbb{R}^{rN \times 3} $ based on the geometric information of the sparse point cloud $ \cX \in \mathbb{R}^{N \times 3} $.
	A well-generated high-resolution (HR) point cloud $ \cY $ should not simply have increased number of points, but should have uniformly distributed points to have similar edge surface to the ground truth.
	Since this upsampling process corresponds to a very challenging ill-posed problem, in order to solve it with a deep learning method, the point feature must well represent the geometry of the low-resolution (LR) point cloud $ \cX $.
	Unlike the existing upsampling method that extracts point features by applying MLP to LR point cloud $ \cX $, we calculate the point-to-point score using the multi-head self-attention proposed by transformer~\cite{vaswani2017attention} with the point feature to make the point cloud feature represent the global structure of the point cloud well.
	However, the point features extracted using transformer do not have the local geometric structure information of the point cloud.
	To solve this problem, we applied EdgeConv proposed in DGCNN~\cite{wang2019dynamic} to multi-head self-attention operation.
	EdgeConv composes the local neighborhood graph of the point cloud, performs a convolution operation on it, and aggregates it so that the point feature represents the geometrical relationship between points in the point cloud.
	As a result, the proposed EdgeFormer calculates the entire relationship of points through multi-head self-attention, so that the global structure of $ \cX $ can be reflected in the point features. Also, it is designed to reflect the local geometric information of $ \cX $ to point features by applying EdgeConv.
	
	\subsection{Network Architecture}\label{sec:network}
	The proposed point cloud upsampling EdgeFormer (PU-EdgeFormer) consists of: i)~encoder, ii)~feature extension, and iii)~coordinate reconstruction as shown in Fig.~\ref{fig:proposed}.
	Each step of PU-EdgeFormer is described as follows:
	
	\textbf{Encoder}
	Since an LR point cloud $ \cX $ is a structure in which the coordinate values $ x, y, z $ of the three-dimensional Euclidean space are simply arranged as row vectors, it is necessary to convert it into a point feature, which is a point set of the latent space.
	In this paper, we construct the module to extract the point feature $ \cF \in \mathbb{R}^{N \times C} $ through MLP operation by concatenating the LR point cloud through four EdgeFormers.
	We will discuss about the EdgeFormer in section~\ref{sec:edgeformer}.

	\begin{figure*}[htb!]
		\centering
		\includegraphics[width=0.99\textwidth]{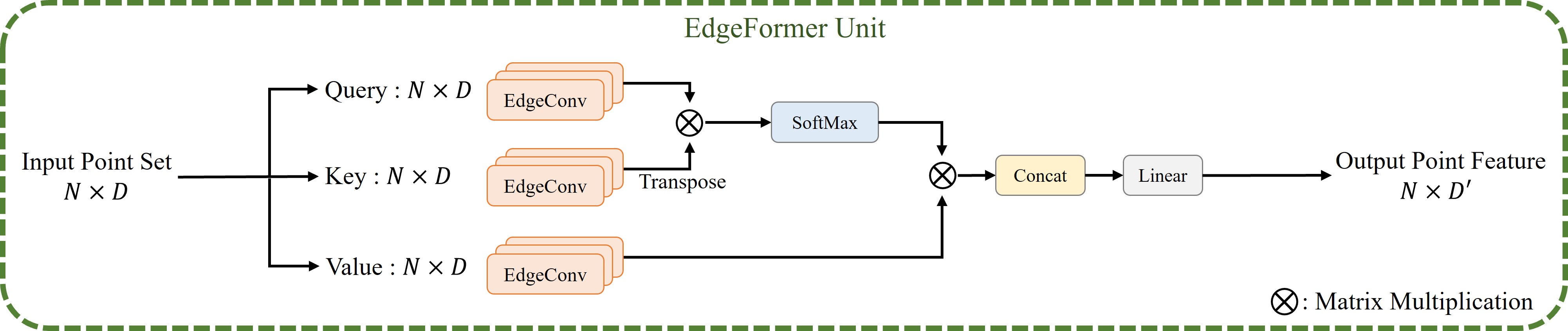}
		\caption{Proposed EdgeFormer unit.}
		\label{fig:edgeformer}
	\end{figure*}
	
    \textbf{Feature Extension}
    To upscale the extracted point feature $\cF$, the feature extension module reshapes $ \cF $ to have a dimension $rN \times C/r$ using a shuffling operator, and then applies the MLP operator, $\Phi_i(\cdot;\theta_i )$, for $ i=1, \cdots, n $.
    A mathematical expression of this process is given as

    \begin{equation}\label{eq:mlp}
    \cF_i = \Phi_i(\cF_{i-1};\theta_i ),
    \end{equation}
    
    \ni where $\theta_i$ denotes a learnable parameter that operated by input point feature $\cF_{i-1}$. 
    In this paper, we set $n=2$, and used 256 and 32 as the output dimension of each MLP. 
    $\cF_n$ in \eqref{eq:mlp} becomes feature extension result $\cF' \in \mathbb{R}^{rN\times C'}$.
    
    \textbf{Coordinate Reconstruction}
	This module generates HR point cloud $\cY$ through the result of feature extension $\cF'$ and LR point cloud $\cX$. The module makes HR point cloud $\cY$ through summation between final Point feature that reshaped by MLP operator $\Phi(\cdot;\theta)$ on $\cF'$ and $\tilde{\cX} \in \mathbb{R}^{rN\times 3}$ that duplicated $ r $ times on $ N $ points:
	
    \begin{equation}
    \mathcal{Y} = \mathcal{\tilde{X}} + \Phi(\cF';\theta).
    \end{equation}
    
    As a result, point feature $\cF'$ that added with duplicated LR point cloud $ \tilde{\cX} $ are important to upsample point cloud, where $\cF'$ is a point set of latent space obtained by the encoder module. 
    Because good upsampling results can be derived when the geometry of the LR point cloud is well reflected as a feature, the encoder plays a very important role in the point cloud upsampling work.

	\subsection{EdgeFormer}\label{sec:edgeformer}
	The encoder shown in the bottom of Fig.~\ref{fig:proposed} consists of four EdgeFormer units.
	An EdgeFormer unit is shown in Fig.~\ref{fig:edgeformer}.

	Unlike an MLP applied to the existing 2D image transformer, in the point cloud, each row vector appears as a point, which is permutation invariance. So the input point set becomes a query, key, and value directly without going through the positional encoding process.
	The EdgeConv operation is applied to query, key, and value, and the results are divided by the number of heads $ h $.
	The query and key divided into each head are matrix multiplied, and the attention score for the current point set is calculated through SoftMax operation.
	By multiplication of this value and matrix, point features are derived in the direction where the value of attention score is high.
	This operation is repeated for each head and all the calculated results are concatenated. Finally, the output point feature is derived through linear operation.
	
	The proposed EdgeFormer replaces the linear operation for query, key, and value with EdgeConv.
	After grouping points using $ k $-nearest neighbor ($ k $-NN) operation, $ \mathcal{N}(\cdot; k) $, EdgeConv performs convolution, $ \psi(\cdot; \theta_f) $, of the grouping results and the learnable parameter $ \theta_f $, and then aggregates the results using max pooling.
	
	\begin{equation}\label{eq:f'}
		f' = \text{MaxPool}(\psi(\mathcal{N}(f; k);\theta_f)),
	\end{equation}
	
	\ni where $f$ denotes each input of query, key, and value, and $f'$ the result of EdgeConv.
	The EdgeConv operation in \eqref{eq:f'} can solve the problem of losing local features when MLP is used.
	We designed the EdgeFormer unit to not only represent the local geometric structure of the input by applying the EdgeConv operation to each input, but also represent the global structure through the multi-head self-attention operation.

	\section{Experiments}\label{sec:experiments}
	
	\subsection{Implementation Details}\label{sec:implementation}

	\textbf{Datasets.}
	We used PU1K dataset that includes large-scale objects with complex shapes provided with PU-GCN~\cite{qian2021pu}.
	The PU1K dataset was collected by PU-GAN~\cite{li2019pu} and ShapeNetCore~\cite{chang2015shapenet}.
	We set the number of points in the LR point cloud of the training dataset to 256 and the number of points in the ground truth to 1024.
	We set the upsampling ratio as $\times4$ just like the other existing methods.
	
	\textbf{Experimental Settings.}
	Our experiments were run on the tensorflow platform.
	As hyper-parameters of the experiment, the batch size was set to 64, the training epochs were set to 100, the learning rate of Adam optimization was set to 0.001, $k = 16$ of EdgeConv in the EdgeFormer unit, and the number of heads $h$ was set to 8. 
	In addition, we performed data augmentation such as rotation, scaling, random perturbations on the training data to avoid overfitting.
	
	\textbf{Loss Functions}
	To train our network, we used Chamfer distance~\cite{yu2018pu} that computes the distance between the two nearest points in two point sets:
	
    \begin{equation}
    \begin{split}
    \selectfont \mathcal{L}(\mathcal{Y}^{Pred}, \mathcal{Y}^{GT}) = {1\over \left\vert \mathcal{Y}^{Pred} \right\vert} \sum_{p \in \mathcal{Y}^{Pred}} \min_{q \in \mathcal{Y}^{GT}} \lVert p - q \rVert_2^2 \\
     + {1\over \left\vert \mathcal{Y}^{GT} \right\vert} \sum_{p \in \mathcal{Y}^{GT}} \min_{q \in \mathcal{Y}^{Pred}} \lVert p - q \rVert_2^2.
    \end{split}
    \end{equation}
	
	\textbf{Evaluation Metrics.}
	According to the quantitative experimental results of recent papers on point cloud upsampling, we use three metrics: Chamfer distance (CD)~\cite{yu2018pu}, Hausdorff distance (HD)~\cite{berger2013benchmark}, and point-to-surface distance (P2F)~\cite{low2004linear}.
	For all of these metrics, lower values indicate better performance.	
	
	\begin{figure*}[htb!]
		\centering
		\includegraphics[width=0.99\textwidth]{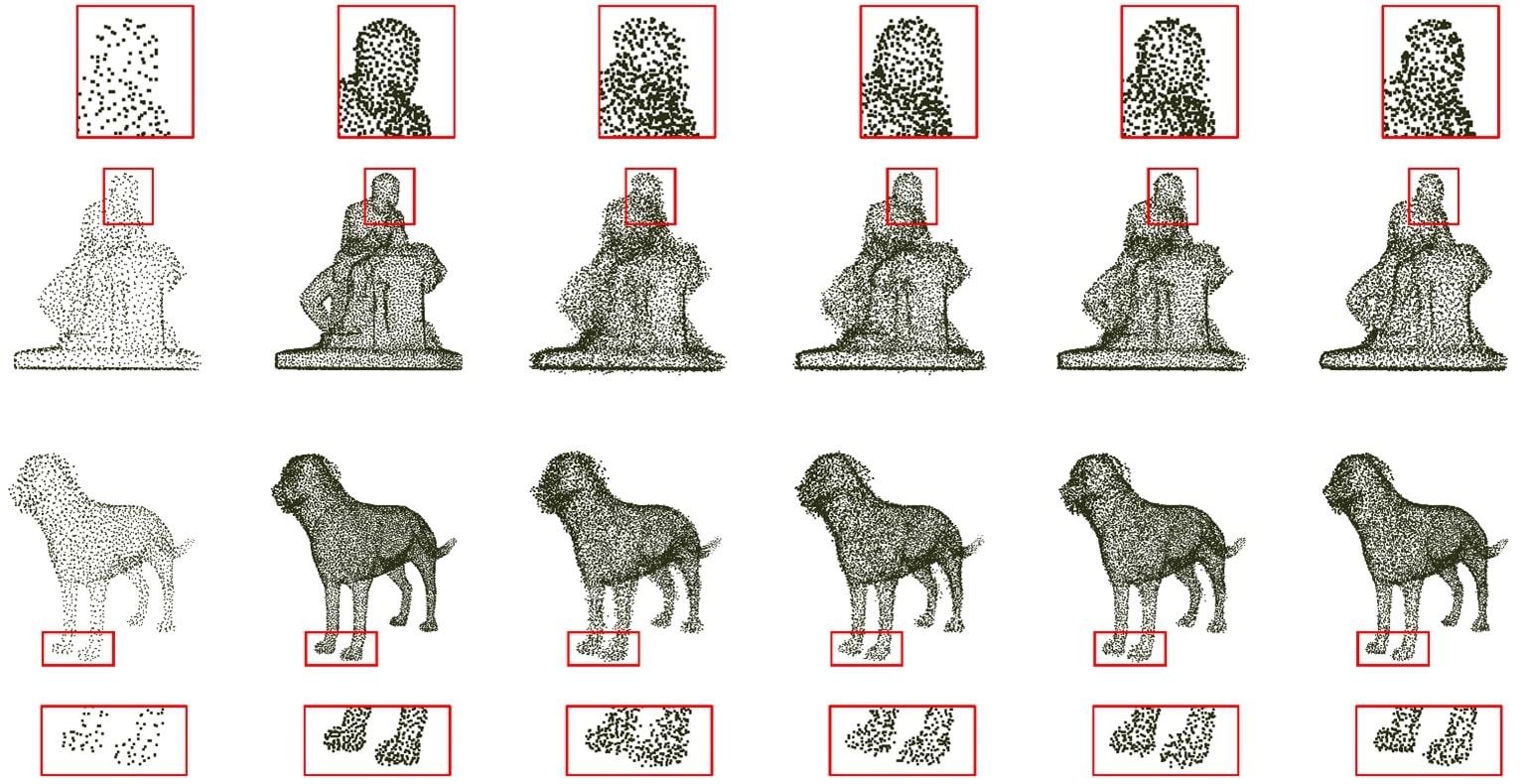}
		\begin{subfigure}{0.16\textwidth}
			\caption{Input}
			\label{fig:Input}
		\end{subfigure}
		\hfill
		\begin{subfigure}{0.16\textwidth}
			\caption{GT}
			\label{fig:GT}
		\end{subfigure}
		\hfill
		\begin{subfigure}{0.16\textwidth}
			\caption{PU-Net}
			\label{fig:PU-Net}
		\end{subfigure}
		\hfill
		\begin{subfigure}{0.16\textwidth}
			\caption{MPU}
			\label{fig:MPU}
		\end{subfigure}
		\hfill
		\begin{subfigure}{0.16\textwidth}
			\caption{PU-GCN}
			\label{fig:PU-GCN}
		\end{subfigure}
		\hfill
		\begin{subfigure}{0.16\textwidth}
			\caption{Ours}
			\label{fig:Ours}
		\end{subfigure}
		\caption{Qualitative upsampling results. We apply x4 upsampling in (a) input that consists of 2048 points. For precise comparison with other methods, we zoom-in on local parts of point clouds in red boxes.}
		\label{fig:comparison}
	\end{figure*}	
	
	\subsection{Comparative Evaluation}\label{sec:comparative}
	
	\textbf{Qualitative Results.} Fig.~\ref{fig:comparison} shows the subjective quality of four different methods including ours. 
	Compared with the existing methods, we show that our method derives less noise and generates HR point clouds with robust edges.
	
	\textbf{Quantitative Results.} 
	Table~\ref{tbl:quantitative} compares the quantitative experimental results.
	Our method shows good performance in CD and HD, similar to state-of-the-art, but shows poor performance in P2F. This is because the edge is excessively smoothed by applying EdgeConv every time in EdgeFormer of Encoder.
	\begin{table}[htb!]
		\caption{Quantitative upsampling results. ↓ means lower value denotes better performance.}
		\label{tbl:quantitative}
		\centering
		\resizebox{\columnwidth}{!}{%
			\begin{tabular}{@{}cccc@{}}
				\toprule
				Methods & CD↓ ($ \times 10^{-3} $)            & HD↓ ($ \times 10^{-3} $)            & P2F↓ ($ \times 10^{-3} $)           \\ \midrule
				PU-Net  & 1.155          & 15.170         & 4.834          \\
				MPU     & 0.935          & 13.327         & 3.551          \\
				PU-GCN  & 0.585          & 7.577          & \textbf{2.499} \\
				Ours    & \textbf{0.462} & \textbf{3.813} & 2.869          \\ \bottomrule
			\end{tabular}%
		}
	\end{table}
	\subsection{Robustness Test}\label{sec:robustness}
	\begin{figure}[htb!]
		\includegraphics[width=0.99\columnwidth]{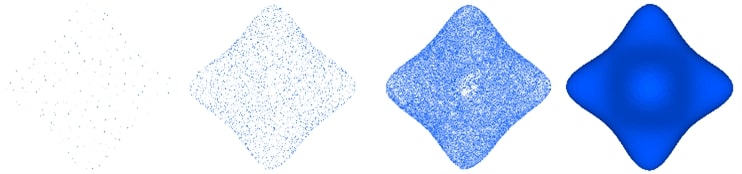}
		\centering
		\begin{subfigure}{0.24\columnwidth}
			\caption{Input}
			\label{fig:Input}
		\end{subfigure}
		\hfill
		\begin{subfigure}{0.24\columnwidth}
			\caption{$\times$ 16}
			\label{fig:GT}
		\end{subfigure}
		\hfill
		\begin{subfigure}{0.24\columnwidth}
			\caption{$\times$ 256}
			\label{fig:PU-Net}
		\end{subfigure}
		\hfill
		\begin{subfigure}{0.24\columnwidth}
			\caption{GT}
			\label{fig:PU-Net}
		\end{subfigure}
		\caption{Sparsity robustness test.}
		\label{fig:sparsity}
	\end{figure}	
	A robustness test was performed to show that our method is robust to sparse and noisy inputs.
	\begin{table}[htb!]
		\centering
		\caption{Quantitative results on noisy input. All values denote CD ($\times 10^{-3}$).}
		\label{tbl:noise}
		\resizebox{0.8\columnwidth}{!}{%
			\begin{tabular}{@{}ccccc@{}}
				\toprule
				Methods & $ \sigma=0.1 $     & $ \sigma=0.5 $     & $ \sigma=1 $               & $ \sigma=2 $           \\ \midrule
				PU-Net  & 1.124          & 1.054         & 1.768         & 3.901 \\
				MPU     & 0.954          & 0.911         & 1.464         & 3.539 \\
				PU-GCN  & 0.632          & 0.809         & 1.416         & 3.410 \\
				Ours    & \textbf{0.589} & \textbf{0.720} & \textbf{1.202} & \textbf{2.880} \\ \bottomrule
			\end{tabular}%
		}
	\end{table}
	
	\textbf{Sparsity Robustness Test.}
	For a sparse input consisting of 256 points, we use upsampling 16 times and 256 times. 
	Fig.~\ref{fig:sparsity} is the visual experimental result, and it can be seen that our method predicts the GT manifold well even when an input is sparse.
	
	\textbf{Noise Robustness Test.}
	We conducted a robustness test by setting the Gaussian noise with standard deviation of 0.1, 0.5, 1, and 2 on the test dataset. Table~\ref{tbl:noise} demonstrates that our method yields robust results against noisy inputs.

	\section{Conclusion}\label{sec:conclusion}
	We proposed EdgeFormer, a new method for extracting features of point clouds.
	EdgeFormer allows the feature to have a global structure through transformer and a local geometric structure through EdgeConv instead of the MLP used when extracting features from the existing point cloud upsampling.
	In the experimental results, PU-EdgeFormer showed higher performance than the existing state-of-the-art, but the edge was excessively smoothed.
	To solve this problem, we will study the method of estimating the manifold in the point cloud and refine it based on this in the future.

	\vfill\pagebreak
	
	\bibliographystyle{styles/ieee}
	\bibliography{references}
	
\end{document}